# Improving Siamese Based Trackers with Light or No Training through Multiple Templates and Temporal Network

Ali Sekhavati*  
Asekh084@uottawa.ca  

WonSook Lee*  
wslee@uottawa.ca  

*School of Electrical Engineering and Computer Science,  
University of Ottawa

**Abstract**

*High computational power and significant time are usually needed to train a deep learning based tracker on large datasets. Depending on available resources, size of datasets and models, training might not always be an option. In this paper, we propose a framework with two ideas on Siamese-based trackers, (i) one is extending number of templates in a way that removes the need to retrain the network and (ii) a lightweight temporal network with a novel architecture focusing on both local and global information that can be used independently from trackers. Most Siamese-based trackers only rely on the first frame as the ground truth for objects and struggle when the target's appearance changes significantly in subsequent frames in presence of similar distractors. Some trackers use multiple templates which mostly rely on constant thresholds to update, or they replace those templates that have low similarity scores only with more similar ones. Unlike previous works, we use adaptive thresholds that update the bag with similar templates as well as those templates which are slightly diverse. Adaptive thresholds also cause an overall improvement over constant ones. In addition, mixing feature maps obtained by each template in the last stage of networks removes the need to retrain trackers. Our proposed lightweight temporal network, CombiNet, learns the path history of different objects using only object coordinates and predicts target's potential location in the next frame. It is tracker independent and applying it on new trackers does not need further training. By implementing these ideas, trackers' performance improved on all datasets tested on, including LaSOT, LaSOT extension, TrackingNet, OTB100, OTB50, UAV123 and UAV20L. Experiments indicate the proposed framework works well with both convolutional and transformer-based trackers. The official python code for this paper will be publicly available upon publication.*

# Introduction

Visual object tracking has numerous applications in various fields, such as security, robotics, traffic control, and sports. With the recent advancements in machine learning, the field of object tracking has greatly improved. Some trackers update the target information during tracking, and some only use the target information in the first frame. Many of the early Siamese based trackers such as SiamRPN++ [3] and SiamFC [1] are examples of the latter. Even though updating target information can help trackers to perform in a more robust way, they can also mistake distractors with the real target and update the target model with such wrong information. This can be quite harmful during tracking, especially when videos include lots of full occlusion and target disappearance for long periods of time. Moreover, in some tracking tasks that are longer and contain many occlusions and target disappearances, wrong updates are more likely to happen and can cause the tracker not to reidentify targets after they reappear. However, not updating the target appearance leads to suboptimal results as well.

Many of the early tracking algorithms were based on modelling the target and updating the model as tracking goes on. The modelling process was performed through various means, such as identifying and utilizing key points or features extracted by a network. Most of the time during tracking, targets undergo small changes, and occlusions occur for only a brief period. However, it is still possible for the target's shape and illumination to change significantly. Therefore, it is crucial to design a model that can adapt easily to changes in the target's appearance. Updating target models is one solution to this problem.

Using target information extracted only from the first frame causes the tracker to have a better reidentification stage. However, it also causes trackers to frequently mistake other objects with the target when the target appearance changes a lot in the middle of videos. An approach to solve this problem, which has been overlooked for years, is making a prediction on target's location. Many trackers, such as SiamRPN++ [3] and TLD [29] look through the whole image after losing the target, without considering where and when it was last seen. This causes them to switch between possible distractors and the target when changes in the target appearance are too much.

Some of the recent trackers, such as Siam R-CNN [5] utilize features obtained from targets in both the first and last frames of videos. In this way, if the target disappears for a long time, it can be reidentified using its appearance in the first frame. Target appearance in the first frame is considered as its ground truth information. This approach is better than using only the target information in the first frame. However, under big changes to the target, this approach must mostly rely on the updates from last frame, which can contain features of a distractor instead.

This paper presents a novel multi-template temporal tracking framework, adaptable to many of the Siamese based trackers to address changes in target appearance during tracking without needing to retrain any network. Moreover, with the use of CombiNet, our proposed temporal network, tracking history is used to prevent trackers from switching between the target and background.

Our approach involves extracting features of the target in the first frame and saving them as the first template in a bag of target templates. This bag contains $n$ templates, and they update considering the tracking confidence score in that frame and a running average over confidence scores when tracking is considered successful. In each frame, a weighted average of the network outputs using each one of those templates is computed. Next, we use a temporal network to predict target location based on its trajectory. In the end, all these are combined and produce a final reliability score.

The reliability score considers 3 things: how close each candidate is to the predicted location, the confidence score of each of those candidates and how robust tracking has recently been. Then the candidate with the highest reliability score is selected as the true target, provided its score is above a certain threshold.

This approach allows for more robust tracking of the targets over time, as the tracker can adapt to changes in appearance by incorporating new templates into the bag of templates. The use of multiple templates also helps trackers to have a better performance when facing occlusions and background clutter. The size of the bag of templates $n$ is a parameter that can be adjusted based on the specific tracking task and the computational resources available. Overall, the proposed approach improves the performance of trackers in various datasets without training on any image.

In order to improve a tracker with this framework, that tracker must be a Siamese based one and produce bounding boxes and scores associated with each of them as outputs. The temporal network proposed in this work can be applied to a broader range of tracking algorithms. To the best of our knowledge, this is the first framework that aims to improve initial results of Siamese based trackers without training on any image.

Overall, our framework makes proper use of target appearance changes and its trajectory history to perform in a more robust way. Experiments on various datasets show the consistent improvement made possible using this framework.

# Related work

In the next part we will have a brief overview of some of the most famous Siamese based trackers and those which use multiple templates.

## Siamese based trackers

It is better to first define what is meant by Siamese based trackers in this paper. All the methods which feed a search image and a template image into a Siamese backbone and then mix these encodings are referred to as Siamese based ones. The backbone can have different architectures, like a convolutional network or a transformer, and the mixing process can be done in different ways, like a cross-correlation operation in SiamFC [1], a Region Proposal Network in SiamRPN [30], or a transformer encoder and decoder, like the one used in SwinTrack [32].

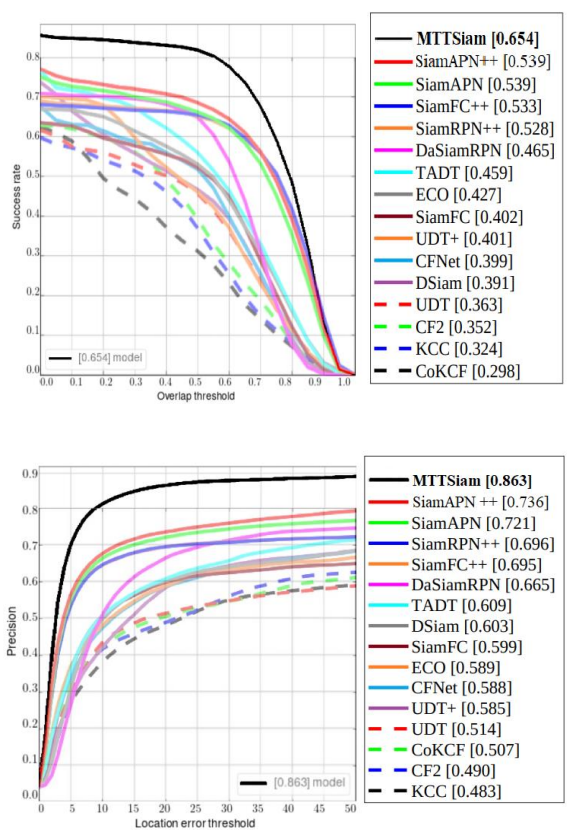

Figure 1: Comparison of our framework when applied on SiamRPN++ over previous state of the art on UAV20L dataset. Constant thresholds were used in this experiment. Our results are denoted with the black line.

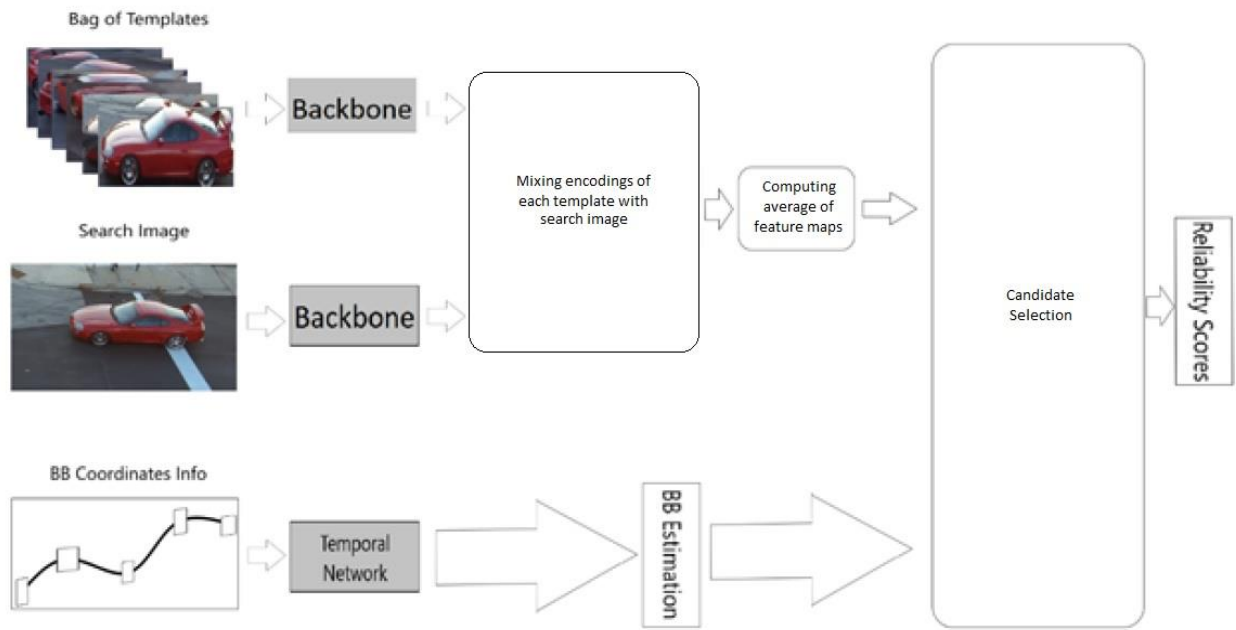

Figure 2: An overview of the pipeline. Given a search image, this algorithm firstly uses *n* templates to identify the target in the search image and computes weighted average of the candidates. The predicted bounding boxes as well as final feature maps including bounding boxes and confidence scores proposed by baseline are given to the candidate selection algorithm to find the most reliable candidate.

The use of Siamese networks for object tracking was first introduced by Bertinetto et al. [1] and has since become widely adopted. They proposed a fully convolutional Siamese network to detect similarities between two images, namely the target in the first frame and the search window in subsequent frames. The target appearance in the first frame served as the anchor image, and the network cross-correlated the feature maps extracted from the convolutional network to locate the target in the search window image of subsequent frames. The simple architecture, accurate results, and high-speed performance of this approach have inspired many researchers to build upon this foundation.

Li et al. [2] added a region proposal network to the Siamese backbone containing a classification and regression branch to the Siamese architecture and highly improved the performance. However, they did not use a very deep network for feature extraction. Deeper networks for extracting meaningful features for tracking was explored by Li et al. [3] in a separate study where they proposed SiamRPN++ tracker by solving the lack of translation invariability to use a deeper network resulting in better accuracy.

One major challenge with object tracking is the issue of mistakenly tracking similar objects to the target. To address this problem, Zhu et al. [4] proposed a solution by modifying the training process and adding a distractor aware module to the original Siamese region proposal network. Another approach was taken by Voigtlaender et al. [5] who designed a two-stage tracker based on the Faster RCNN [7] detection algorithm, followed by dynamic programming to separate the real target from the distractors. However, this method requires heavy computations and does not benefit from target trajectory information.

## Multiple templates

The idea of using multiple templates during tracking has been there for a long time, yet it has been covered only by a few methods. For instance, the MMLT tracker [23] separates its memory model into long-term and short-term sections. Each of those sections contain many templates extracted from different frames. When the tracking score is reliable, both memory stores contribute to tracking. When target reidentification must be performed, it only uses the long-term features. This work has similarities with our work in the sense that they use multiple templates as well. However, they combine the output of each template before performing cross-correlation. In our work, the final outputs are combined to remove the need for training trackers with the new settings and make the framework adaptable with other Siamese-based trackers. Moreover, MMLT uses 2 networks and 100 memory stores for capturing target Siamese and semantic features in 100 frames but does not consider trajectory history of the target. With the help of CombiNet, the lightweight temporal network proposed in this work, this framework makes predictions over where the target can be found. This greatly reduces the effect of distractors that are in different parts of an image.

Figure 3: An example of how the bag of templates gets updated using 6 templates. Template #1 contains the target in the first frame and never gets updated. Other templates from 2 to 6 contain different views of the target and they get replaced by new target appearances. Template 2 contains the most similar features and template 6 is the most diverse one. The orange bordered template shows which template is updated for each frame.

Siam R-CNN [5] uses two templates during tracking. The tracker utilizes the target's last frame features when it has been continuously tracked. However, using only one extra template can be dangerous as it might contain wrong updates. Moreover, by relying on target's appearance in the last frame, the tracker might confidently continue tracking a distractor, which is worse than only losing the target.

Siam R-CNN also uses a constant threshold for updating the template which leads to suboptimal results.

Yan et al [24] proposed a transformer-based tracker with an architecture similar to Siamese networks. They used a backbone to process the search region, target template in the first frame, and the last reliable target image. The backbone output was then flattened and fed into a transformer architecture along with a bounding box prediction head to find the target. While this approach showed improvement over previous trackers, it can be susceptible to false updates and may track distractors instead of the target. To mitigate this problem, our work aims to increase the number of templates used to capture target features at different poses, lighting conditions, etc. By using more templates, the influence of false updates is reduced, which leads to more accurate tracking. The reason is that in case one template updates with wrong information, there are still n-1 correct ones.

## Method

This section provides a more detailed and clear explanation of the proposed framework, Multi-Template Temporal Siamese (MTTSiam) architecture. Figure 2 shows an illustration of this method.

### Multiple templates

Despite recent improvements in Siamese-based tracking, they still face challenges in handling changes that targets might have during tracking. This is mainly due to the fact that most of these trackers rely on a single template extracted from the target's appearance in the first frame. Recognizing the limitations of this approach, we use multiple templates, each containing the target's features extracted by the backbone at a different frame. The first template contains the ground truth information of the target in the first frame, and it remains fixed as a reference for tracking throughout the video sequence.

Templates are updated in such a way that the bag contains templates which are very similar to the target, as well as the ones which are slightly diverse yet belong to it. The reason is that collecting only templates that are too similar does not contribute enough to improving the tracker. However, it might increase the risk of false updates. It is tried to reduce that effect by having *n* templates. Moreover, if *n* is large enough, the tracker will have chances to fix wrong updates by replacing false information with correct ones during tracking. This approach enables our algorithm to store both similar appearances of the target and those that are less similar but belong to the target. Moreover, since updates happen whenever the confidence score is above a minimum threshold, the framework contains recent features of the target as well.

Each template slot in our approach is associated with a threshold for classification score to update. Those thresholds are set adaptively using the average confidence score $\bar{C}$ when confidence score is above $\tau_{min}$. $\tau_{min}$ indicates minimum confidence score for updating the bag of templates. There are two types of templates: those with thresholds above $\bar{C}$ and those with thresholds below it. The threshold for the first group is set using Equation 1:

$$\tau_i = 1 - \frac{(1 - \bar{C})}{T_i} \quad (1)$$

where $T_i$ is the weight threshold for the $i$-th template. The second group update using Equation 2:

$$\tau_i = 1 + \frac{(\tau_{min} - \bar{C})}{T_i} \quad (2)$$

Each template updates when the confidence score $C$ of that frame is between $\tau_i$ and $\tau_{(i-1)}$. $\tau_{(i-1)}$ is always higher than $\tau_i$, and $\tau_1$ is set to 1 since the first template never updates. Figure 3 shows an example of member update in the bag of dynamic templates.

The extracted templates, denoted by $Z_i$, are used in conjunction with the ground truth information represented by $Z_1$. We use the transformation f to extract features from these templates, as shown in Equation (3):

$$f(z_i) = E_i \quad (3)$$

where $i$ represents the template number, and $f$ is the Siamese backbone. The features $E_i$ and the current search region $E_x$ are processed by the second stage of Siamese based tracker, which is mixing information. This step is dependent on the baseline and can vary from tracker to tracker. When using SwinTrack [32] as the baseline, in this step template embeddings are separately combined with search embeddings and then processed by transformer encoder and decoder. After that, the head network produces feature maps. In the end, the final feature maps made using each template are combined through a weighted average as demonstrated by Equation 4:

$$M = \sum_{i=1}^{n} W_i m_i \quad (4)$$

where $W_i$ denotes the specified weight for template $i$, $m_i$ is the feature map produced by the $i$-th template, $M$ is the final feature map and $n$ is the number of templates. The weights are obtained through experiments. Swin Track produces two feature maps in the output, one for classification and one for regression. Using that tracker as the baseline allows us to produce two final feature maps as well. Finally, the top 10 candidates are stored to be processed by the candidate selection algorithm.

## Candidate selection algorithm

The candidate selection algorithm is responsible for combining scores and candidates' coordinates with path information. Its purpose is to prevent the tracker from mistakenly switching between possible similar distractors and the target, and to know whether the tracked object is a distractor or the real target. To this end, the first step is to approximate the position of the target in the current frame, using only its coordinates in the previous frames. In this paper, the prediction is done through a temporal network that uses the bounding box coordinates of the previous 4 frames to predict target position in the next frame.

## Temporal network

A temporal network is trained on the GOT10K dataset to predict the target coordinates using the bounding boxes produced in the previous 4 frames. Since images have different resolutions, we divided the height and width coordinates of bounding boxes by image height and width. In this way, all coordinates remain between 0 and 1 and make the regression task simpler for the network. During training the network, L2 regularization was used to prevent it from overfitting. The network has six layers. In each of the first 4 layers, a convolutional layer and a fully connected layer are applied in parallel to the input. This is because CNNs are more capable of capturing local features and MLPs are better at finding global ones. Finally in that layer, the average of their outputs is computed and reshaped so that it can be processed by the next convolutional and fully connected layer. Since the output of every layer is the combined features captured from a convolutional and fully connected layer, we name our network CombiNet. The last two layers are fully connected ones.

## Reliability score

Having the final similarity scores and an approximation of where the target can be found in the current frame, it is time to combine the two to find the most reliable candidate. For each of the 10 selected candidates, firstly a distance error is computed. This is done by measuring the absolute difference of the predicted coordinates and the given coordinates, as given in equation 5:

$$DE_j = \frac{|PC_1 + PC_2 - CC_{j,1} - CC_{j,2}|}{2} \quad (5)$$

where $PC_1$ and $PC_2$ is the predicted coordinate on *x* and *y* axis, $CC_{j,1}$ and $CC_{j,2}$ is the *j*-th candidate coordinate and $DE_j$ is the distance error for candidate *j*. Finally, the reliability score for candidate *j* is calculated with Equation 6:

$$RS_j = C_j - \frac{(DE_j - b) * SC_t * (1 - C_j)}{RW} \quad (6)$$

where $RS_j$ is the final reliability score for candidate *j*, $SC_t$ is the Sequential Confidence value at frame *t*, *b* (=0.1) is a bonus threshold and *RW* is a weight that determines how much effect path information should have over similarity score. If *DEj* is below *b*, the term on the right turns into a positive value and makes *RSj* to be more than *Cj*. For every frame, *SC* gets updated using a running average when tracking is successful.

When the similarity score remains high for a sequence of frames, the *SC* value increases. Otherwise, it decreases. It can be considered as a value determining how confident the tracker has been in recent previous frames. Equation 6 demonstrates that sequential confidence effectively reduces reliability score if both *SC* and *DE* are high. It also increases that score when sequential confidence is high, and distance error is low.

# Experiments

## Results

According to the experiments on the OTB [8], UAV [9], LaSOT [26], LaSOT extension [28] and TrackingNet [25] datasets, MTTSiam improves the success rate and precision of the baselines applied on. The improvement is due to the use of more templates and the candidate selection algorithm with bounding box prediction, which prevents the tracker from switching between distractors. Table 1 and Table 5 respectively provide more detailed results on the OTB 100 and OTB 50 datasets using SiamRPN++ as the baseline. Table 2, 3 and 4 show the results on LaSOT [26], LaSOT extension [28] and TrackingNet [25] respectively.

**Table 1:** Results on OTB 100 with SiamRPN++ as the baseline. _R denotes using ResNet 50 and _A denotes the use of AlexNet for the backbone.

| Tracker | Success | Precision |
|---|---|---|
| SiamFC [1] | 58.7 | 77.2 |
| SiamFC++ [16] | 68.3 | 91.2 |
| KYS [17] | 69.5 | 91.0 |
| Dimp [14] | 68.6 | 89.9 |
| SiamRPN++ [3] | 69.6 | 91.4 |
| GradNet [18] | 63.9 | 86.1 |
| MDNet [19] | 67.8 | 90.9 |
| Ocean [20] | 67.2 | 90.2 |
| SiamDW [21] | 67.4 | 90.5 |
| ECO [22] | 69.1 | 91.0 |
| MTTSiam_A (ours) | 67.4 | 88.7 |
| MTTSian_R (ours) | 70.4 | 91.6 |

**Table 2:** Results on LaSOT [26].

| Tracker | Success rate | Precision rate |
|---|---|---|
| Siam R-CNN [5] | 64.8 | - |
| TrDiMP [32] | 63.9 | 61.4 |
| TransT [33] | 64.9 | 69.0 |
| STARK [24] | 67.1 | - |
| KeepTrack [34] | 67.1 | 70.2 |
| SwinTrack-B-384 [32] | 71.3 | 76.5 |
| SwinTrack-B-384-MTTSiam (ours) | **72.2** | **77.4** |

Figure 4 shows an example of how well MTTSiam performs in tracking a car with a moving camera where the objects are very blurred when applied on SiamRPN++.

**Table 3:** Results on LaSOT extension [28].

| Tracker | Success rate | Precision rate |
|---|---|---|
| SiamRPN++ [3] | 34.0 | 39.6 |
| AutoMatch [36] | 37.6 | 43.0 |
| DiMP [14] | 39.2 | 45.1 |
| LTMU [35] | 41.4 | 47.3 |
| KeepTrack [34] | 48.2 | - |
| SwinTrack-B-384 [32] | 49.1 | 55.6 |
| SwinTrack-B-384-MTTSiam (ours) | **50.1** | **57.0** |

**Table 4:** Results on TrackingNet [25].

| Tracker | Success rate | Precision rate |
|---|---|---|
| Siam R-CNN [5] | 81.2 | 80.0 |
| TrDiMP [32] | 78.4 | 73.1 |
| TransT [33] | 81.4 | 80.3 |
| STARK [24] | 82.0 | - |
| SiamRPN++ [3] | 73.3 | 69.4 |
| SwinTrack-B-384 [32] | 84.0 | 82.8 |
| SwinTrack-B-384-MTTSiam (ours) | **84.1** | **83.5** |

## Datasets

In this section we briefly introduce the datasets that were used for evaluating this work and how MTTSiam performed on them.

UAV123: It is a dataset containing 123 high-resolution videos captured by unmanned aerial vehicles (UAV) or a simulator using the Unreal Engine 4 (UE4). There are more than 110 K frames in this dataset. There are 3 types of videos in this dataset: 103 videos are taken from professional UAVs, 12 videos captured by low-cost UAVs and 8 videos captured by a UAV simulator. We tested this framework using SiamRPN++ as the baseline. MTTSiam shows a lot of improvement over the baseline. This might be due to more predictable changes in target trajectory when looking from above.

UAV20L: Our experiments on the UAV20L [9] long-term tracking dataset suggests that having multiple templates of the target and relying on its path information can greatly enhance tracker performance. Figure 1 and table 6 compare MTTSiam with other state-of-the-art trackers on the UAV20L [9] dataset. This dataset contains 20 long high-resolution videos. It is a subset of the UAV123 dataset.

OTB: OTB100 dataset contains 100 videos taken from different viewpoints. Videos in this dataset are low-resolution and trackers are evaluated using the success and precision plot of their proposed bounding boxes. OTB50 is a small subset of this dataset containing 50 sequences and is one of the oldest datasets in object tracking.

LaSOT: The dataset contains 1400 sequences and 280 of them are for validation. The baseline used for this dataset is SwinTrack and MTTSiam improves both success rate and precision.

LaSOT extension: It is made of 150 challenging videos. MTTSiam improves SwinTrack as it did on LaSOT dataset as well.

TrackingNet: This is the first large-scale dataset containing videos for tracking in the wild. MTTSiam makes slight improvements over the baseline.

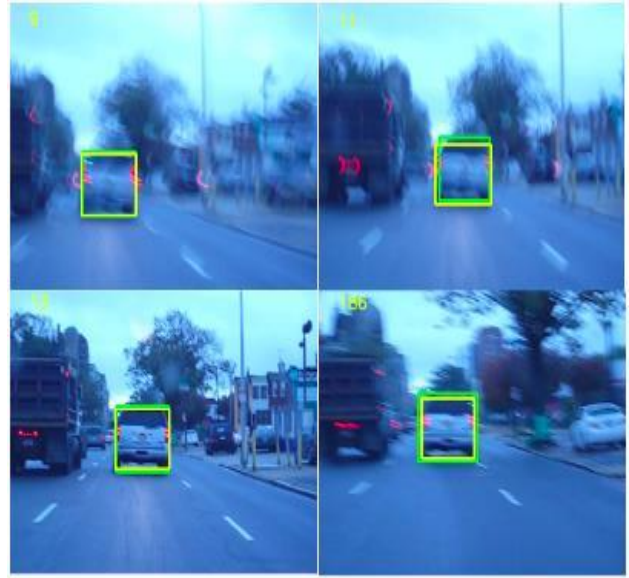

Figure 4: MTTSiam performance in low resolution blurry images. The green line shows the ground truth and the yellow line is proposed by MTTSiam. Images are from the OTB100 dataset [8].

**Table 5:** Results on OTB 50. _R denotes using ResNet 50 as the backbone and _A denotes the use of AlexNet for the backbone. MTSiam means using different templates without the robustifier and SiamR means using the temporal path predictor network without using multiple templates.

| Tracker | SiamRPN++_A | SiamRPN++_R | MTSiam_A | MTTSiam_R | MTTSiam_A | SiamR_A |
|---|---|---|---|---|---|---|
| Success | 61.7 | 66.5 | 64.2 | **66.8** | 63.3 | 61.9 |
| Precision | 83.3 | 88.6 | 87.6 | **89.5** | 85.7 | 83.9 |

**Table 6:** Results of comparison on UAV20L [9] long-term tracking dataset. Our method shows a lot better performance by a large margin.

| Tracker | SiamFC [1] | HiFT [10] | DaSiamRPN [4] | ECO [22] | SiamRPN++ [3] | MTTSiam (ours) |
|---|---|---|---|---|---|---|
| Success ↑ | 40.2 | 56.6 | 46.5 | 59.9 | 52.8 | **65.4** |
| Precision ↑ | 59.9 | 76.3 | 66.5 | 42.7 | 69.6 | **86.3** |

## Ablation study

Table 5 and 7 show the effect of each of our contributions. According to our experiments, the candidate selection algorithm plays a bigger role when the templates are being replaced as it helps to choose better templates. It also shows that when multiple templates are used and the candidate selection algorithm chooses the most reliable candidate, it achieves the best results. Table 7 contains some of our experiments on the UAV and OTB datasets.

**Table 7:** Ablation study on 3 datasets. The numbers show the success rates. Different thresholds were used for each dataset when not using adaptive type.

| Tracker | OTB 100 | UAV123 | UAV20L |
|---|---|---|---|
| SiamRPN++ ResNet50 | 69.6 | 61.3 | 52.8 |
| 6 templates AlexNet | 67.4 | 58.1 | - |
| 6 templates ResNet50 | 70.4 | 62.2 | 65.3 |
| 10 templates ResNet50 | 70.4 | 63.3 | 65.4 |
| adaptive thresholds ResNet50 | 70.1 | 62.8 | 64.5 |

### Number of templates

In our experiments, we set the number of templates to 6 and 10. Increasing the number of templates from 1 to 6 hugely improves the results on all datasets. However, increasing them from 6 to 10 slightly improves the results.

### Effect of different backbones

We tried two different backbones with SiamRPN++: AlexNet and ResNet 50. The success rates obtained by the Resnet 50 in all the experiments are higher than those obtained with AlexNet. However, when using 10 templates, MTTSiam with AlexNet as backbone runs 3.2 times faster than when ResNet 50 is used (58 fps vs 18 fps using a single NVIDIA RTX 3080 GPU, and 3 fps vs 0.5 fps using only an intel core i5 CPU). With SwinTrack only the main backbone was tested.

### Thresholds for updating templates

We tried both adaptive thresholds and constant ones. Constant thresholds make the tracker perform better on some datasets, but the same values do not help in others, and sometimes might even have bad influence. This depends on many factors such as dataset difficulty, image resolution, etc. When using a different backbone, the same constant thresholds do not help either. Adaptive thresholds on the other hand always make improvements without the need to change any value for different datasets. For fair comparison, the results reported in this paper use adaptive thresholds without any changes when applying on different datasets.

### Training details

For training the temporal network, we used the GOT10k dataset. Having the bounding box coordinates of every 4 sequential frames, the regressor network predicts the bounding box coordinates of the target in the next one. A batch size of 16384 was used and the network was trained with SGD optimizer with momentum value set to 0.9 and weight decay set to 0.001 for 1000 epochs. Learning rate is 0.4 at the beginning, and after each epoch it is replaced by $0.99^{epoch\ number}$. For finding the similarities, the baselines are not retrained and the pretrained models provided by them are used. The official python code will be available on github.

## Conclusion

In this paper a new framework for improving Siamese based trackers is introduced. This work requires minimal training for CombiNet and when it is trained, applying the framework on other trackers does not need further training. By using target features extracted from different frames and predicting its potential position using its path history, it improves different baselines. The huge improvement on UAV20L [9] dataset suggests that this framework might perform better with long-term datasets and trackers.